\documentclass[sigconf]{acmart}
\usepackage{multirow}
\usepackage{enumitem}
\usepackage{subcaption}
\usepackage{bbm}
\usepackage{graphicx}
\usepackage{mathtools}
\usepackage[linesnumbered,ruled,vlined]{algorithm2e}
\usepackage{multirow}
\usepackage{bbm}
\usepackage{balance}

\newcommand{\modelname}{\textsf{ELECRec}\xspace}

\AtBeginDocument{%
  \providecommand\BibTeX{{%
    \normalfont B\kern-0.5em{\scshape i\kern-0.25em b}\kern-0.8em\TeX}}}

\copyrightyear{2022}
\acmYear{2022}
\setcopyright{acmcopyright}\acmConference[SIGIR '22]{Proceedings of the 45th
International ACM SIGIR Conference on Research and Development in Information
Retrieval}{July 11--15, 2022}{Madrid, Spain}
\acmBooktitle{Proceedings of the 45th International ACM SIGIR Conference on
Research and Development in Information Retrieval (SIGIR '22), July 11--15, 2022,
Madrid, Spain}
\acmPrice{15.00}
\acmDOI{10.1145/3477495.3531894}
\acmISBN{978-1-4503-8732-3/22/07}

\setlist[itemize]{leftmargin=*}
 
\begin{document}

\fancyhead{}

\title{
\modelname: Training Sequential Recommenders as Discriminators
}

\author{Yongjun Chen}
\affiliation{%
  \institution{Salesforce Research}
\country{Palo Alto, CA, USA}
}
\email{yongjun.chen@salesforce.com}

\author{Jia Li}
\affiliation{%
  \institution{Salesforce Research}
  \country{Palo Alto, CA, USA}
  }

\email{jia.li@salesforce.com}

\author{Caiming Xiong}
\affiliation{%
  \institution{Salesforce Research}
  \country{Palo Alto, CA, USA}}
\email{cxiong@salesforce.com}

\begin{abstract}
Sequential recommendation is often considered as a generative task, 
i.e., training a sequential encoder to generate the next item 
of a user’s interests based on her historical interacted items. 
Despite their prevalence, these methods usually 
require training with more meaningful samples to be effective, 
which otherwise will lead to a poorly trained model. In this work, 
we propose to train the sequential recommenders as discriminators 
rather than generators. Instead of predicting the next item, 
our method trains a discriminator to distinguish if a sampled item 
is a ‘real’ target item or not. A generator, as an auxiliary model, 
is trained jointly with the discriminator to 
sample plausible alternative next items and will be thrown out after training.
The trained discriminator is considered as the final
SR model and denoted as \modelname.
Experiments conducted on four datasets
demonstrate the effectiveness and efficiency of the 
proposed approach
\footnote{Code is available at
\href{https://github.com/YChen1993/ELECRec}{https://github.com/YChen1993/ELECRec}}.
\end{abstract}


\maketitle

\section{Introduction}

Recommender systems play an essential role in 
many applications~\cite{dewet2019finding,le2019correlation}, including e-commerce,
advertising, and grocery shopping, to 
alleviate information overload to users.
Sequential recommendation~\cite{rendle2010factorizing,quadrana2017personalizing,kang2018self,sun2019bert4rec,ma2020disentangled} is one of the core tasks
in recommender systems, aiming to capture the 
sequential dynamic of users' behaviors from their
historical behavior sequences.

\begin{figure*}[htb]
  \centering
  \includegraphics[width=0.85\linewidth]{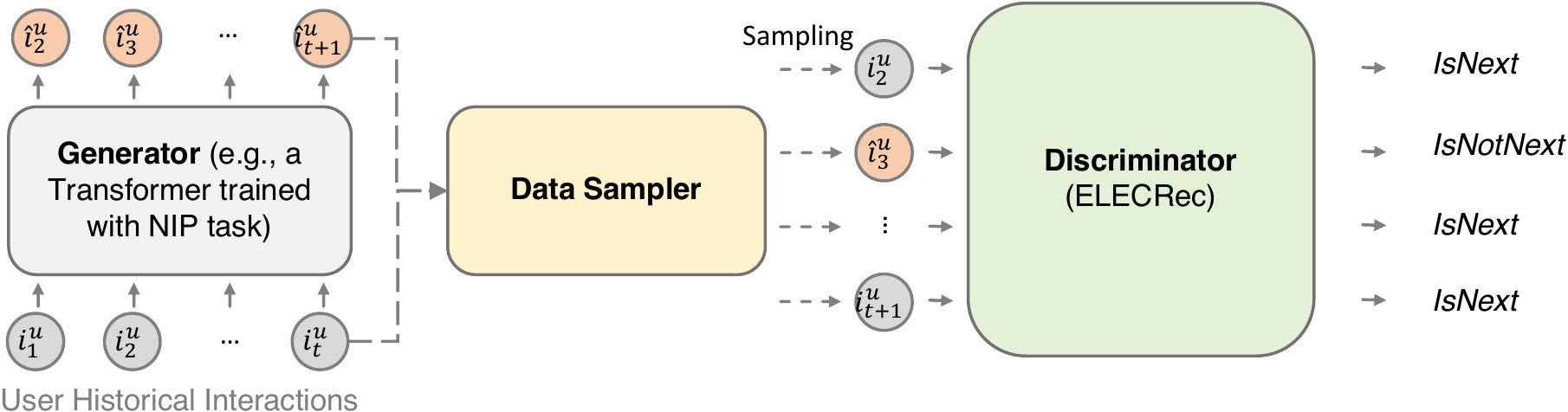}
  \caption{Illustration of the Proposed Learning Framework. 
  The generator takes user behavior sequences as inputs to perform
  next-item prediction (NIP) task. The plausible items are sampled
  from the generator and combined with the original sequences to
  form the new inputs for discriminator. The discriminator
  aims at distinguishing that if an sampled item is the target item or not.
 }
  \label{fig:illustration}
\end{figure*}

Recent works~\cite{he2016fusing,kang2018self,li2020time,chen2021modeling,fan2022sequential,chen2022intent} commonly utilize
Transformers~\cite{vaswani2017attention,radford2019language} to encode sequence 
and train the sequential recommender
via a generative task: next-item prediction (NIP), i.e., 
take a user's historical
behavior sequence as input and then train the model 
to predict the next item. 
While Transformer itself has shown effective modeling sequential dynamics of item correlations~\cite{kang2018self,zhang2019feature,zhou2020s3,ma2020disentangled,li2021lightweight}, 
training via such generative (NIP) task can 
lead to a poorly trained model without enough representative training samples.
For example, given a user behavior sequence ``Jacket, Shoes, Pants, Sweater, Basketball'',
the model can often fail to predict the next item as ``Basketball'' after the user purchases Sweater.
Because purchasing ``Basketball'' is a rare consumer behavior right after purchasing a series of apparel.
Without enough representative sequences such as ``Shoes, Jacket, Sweater, Baseball'', ``Hat, Jacket, Shoes, Football'', etc.,
the model cannot capture the sequential correlations between apparels and
sporting goods, resulting in a poorly trained SR model.

Alleviating the issue mentioned above under the generative task
is challenging because it requires more meaningful training data
while recommender systems often facing cold-start and data-sparsity issues~\cite{zhu2021transfer,wang2021sequential,liu2021contrastive,wu2021self}.
S$^3\text{-Rec}$~\cite{zhou2020s3} and CL4SRec~\cite{xie2020contrastive} propose to
randomly ``reorder'', ``mask'',
and ``crop'' sequence as the augmented user behavior sequence to enrich the training set.
However, if a user behavior sequence is sensitive to the random permutations,
these augmentations will bring additional noise into the training samples.
As a result, the learned item correlations can be more inaccurate.
ASReP~\cite{liu2021augmenting} and BiCAT~\cite{jiang2021sequential}
propose to train unidirectional 
and bidirectional Transformers with reversed sequences 
to generate pseudo-prior items, respectively.
Although they alleviate the user cold-start issue~\cite{zhu2021transfer,wang2021sequential}, the aforementioned issue can still presence,
i.e., the generated pseudo-prior items via Transformer are most likely did not 
include sporting goods because the Transformer is also trained via NIP task.
Besides, training via NIP often requires
more iterations
to
learn the accurate user preference to items well, 
which suffers a substantial computation cost.
Adersarial training
via generative adversarial networks (GANs)~\cite{goodfellow2014generative,wang2017irgan}
is challenging because of the difficulty 
of back-propagating from the adversarial networks to the generator. 
The training process is also not stable, leading to a worse-performed model.

Inspired by the success of ELECTRA~\cite{clark2020electra} in efficient
language model pre-training, we propose to train the sequential recommenders
as a discriminative task rather than a generative task.
The discriminater is trained to identify the `real' next item
from the plausible alternatives. 
To generate these plausible items, 
we jointly train a generator, 
an auxiliary model trained with the NIP task. 
Then a certain percentage of items in the original sequence 
is replaced with the items generated from the generator.
Our training strategy is more effective because it generates high-quality
plausible samples from a learnable generator rather than manually crafting, 
making the discriminator steadily lift its ability to discriminate the true item correlations.
Treating the task as a binary classification task is also more efficient. 
It only selectively updates item embeddings in the sequences 
without computing all items required in NIP. 
We denote the trained discriminator as \modelname for
\textbf{E}ficiently \textbf{L}earning an \textbf{E}ncoder as a dis\textbf{C}riminater 
for sequential \textbf{Rec}ommendation.
Different from ELECTRA in NLP domain
that distinguishes if a word token at ``current'' position 
is generated or from the training data for text understanding, our method aims at
distinguishing the plausible ``future'' items, which is better aligned with
the recommendation tasks.

We conduct extensive experiments on four datasets. 
The empirical results demonstrate the effectiveness 
and efficiency of our proposed framework 
for training a sequential recommendation model. 
The detailed analysis also shows the superiority
of \modelname. We summarize 
our contributions as follows:

\begin{itemize}
    \item We propose a novel training framework,
    which alternatively trains a sequential recommender as a discriminator rather
    than a generator. 
    \item We propose to generate plausible alternative items via a jointly trained 
    generator and suggest sharing the item embeddings between generator and discriminator networks
    to boost efficiency.
    \item We empirically demonstrate the effectiveness and efficiency of \modelname,
    which is trained via our proposed training framework.
\end{itemize}

\section{Preliminary}

\subsection{Problem Statement}
We denote a set of users and items as $U$ and $V$, respectively. 
Each user $u \in U$ is associated with a sequence of interacted items
that are 
sorted in chronological order:
$I^{u} = \{i^{u}_{1}, i^{u}_{2}, \cdots, i^{u}_{n_{u}}\}$.
Specifically,
$i_{n_{u}}$ is the item that user $u$ is interacted at $n_{u}$ step.
The goal of \emph{Sequential Recommendation} (SR) is to accurately predict
the next item $i_{n+1}$ that user likely to interact with given $I^{u}$.

\subsection{Next-Item Prediction}

Existing works~\cite{hidasi2015session,tang2018personalized,li2020time,zhang2019feature,zhou2020s3} 
commonly consider the aforementioned problem as a generative task
to recover the \emph{next} user interacted item $i^{u}_{t+1}$ based on 
the truncated (with the ``padding'' operator)~\cite{hidasi2015session,tang2018personalized,kang2018self} 
prior interactions $\{i^{u}_{1}, i^{u}_{2}, \cdots, i^{u}_{t}\}$ during training stage.
The common training loss to achieve that goal is this kind:

\begin{equation}
\label{eq:next-item-all}
\mathcal{L}_{\mathrm{NIP}} = \sum_{u=1}^{N} \sum_{t=2}^{T} \mathcal{L}_{\mathrm{NIP}}(u,t),
\end{equation}

\begin{equation}
\label{eq:next-item-sing}
\mathcal{L}_{\mathrm{NIP}}(u,t) = -\log p_{\theta}(i^{u}_{t+1}|i^{u}_{1}, i^{u}_{2}, \cdots, i^{u}_{t}),
\end{equation}
where $\theta$ describes a neural network $f_{\theta}$, which encodes sequence latent representation space:
$\mathbf{h}_{t}^{u} = f_{\theta}(\{i^{u}_{j}\}_{j=1}^{t})$ and  the probability is
often defined to measure the similarity between the encoded sequence $h_{t}^{u}$
and the next item $i^{u}_{j+1}$ in the representation space.

\section{METHODOLOGY}
In this section, we present a novel training framework, 
which contains a \emph{generator} $G$, a \emph{data sampler},
and a \emph{discriminator} $D$. The sequence encoders that used in both $G$
and $D$ are Transformer networks~\cite{vaswani2017attention,radford2019language} (denoted as $f_{G}(\cdot)$ and $f_{D}(\cdot)$),
which are widely used on modern SR models~\cite{kang2018self,zhang2019feature,zhou2020s3,ma2020disentangled,li2021lightweight}.
Figure~\ref{fig:illustration} illustrates the overall learning framework.

\subsection{Generator}
The role of generator $G$ is to generate plausible next items to 
improve the discrimination ability
of discriminator $D$.
To achieve that, $G$ is trained with the NIP task (Eq.~\ref{eq:next-item-all}). 
At each time $t$, the probability of an item is considered to be generated defined with a Softmax layer:

\begin{equation}
\label{eq:generator}
p_{G}(i^{u}_{t+1}|\mathbf{h}_{t}^{u}) = \exp((\mathbf{i}_{t+1}^{u})^{T}\mathbf{h}_{t}^{u})/\sum_{i' \in V} \exp ((\mathbf{i'})^{T}\mathbf{h}_{t}^{u}),
\end{equation}
where $\mathbf{h}_{t}^{u} = f_{G}(\{i^{u}_{j}\}_{j=1}^{t})$ and $\mathbf{i'}$ denotes the embedding of item $i'\in V$.

\subsection{Data Sampler}
For a given sequence $I^{u}$, the data sampler samples $\alpha$ percentage of items
from the generator $G$ to replace the original target items.
Formally, we first sample total $\text{ceil}(\alpha \cdot T)$ positions of the sequence: $\{r_{j}\}_{j=1}^{\text{ceil}(\alpha \cdot T)} \sim \text{uniform}\{1, T \}$, where $ j \in (1, \text{ceil}(\alpha \cdot T))$. Then we sample the items from the generator $G$ for every sampled position $r_{j}$ as follows:
$\hat{i}_{j}^{u} \sim p_{G} (i_{j}^{u}|\mathbf{h}_{j-1}^{u})$ for $j \in r$. Finally, we replace the original target items
with the sampled items for each position $r_{j}$ to form the new input sequence $\hat{I}^{u}$ for discriminator $G$.

\subsection{Discriminator}
For a given input sequence $\hat{I}^{u}$, the discriminator aims at predicting that whether 
$\hat{i}^{u}_{t}$ is a ``real'' or ``fake'' target item.
Notes that if the sampled items from the generator are indeed the
target items, they are considered as target items instead of ``fake'' ones.
Formally speaking, the discriminator $D$ distinguish items with a sigmoid function:
\begin{equation}
\label{eq:discriminator}
p_{D}(\hat{\mathbf{h}}_{t}^{u}) = \text{sigmoid}(\mathbf{w}^{T}\hat{\mathbf{h}}_{t}^{u}),
\end{equation}
where $\mathbf{w}$ is a learnable parameter and  $\hat{\mathbf{h}}_{t}^{u} = f_{D}(\{\hat{i}^{u}_{j}\}_{j=1}^{t})$.

The discriminator is trained with the binary cross-entropy loss as follow:

\begin{equation}
\label{eq:dis-bce}
\begin{split}
\mathcal{L}_{Disc} = \sum_{u=1}^{N} \sum_{t=2}^{T} 
&- \mathbbm{1}(\hat{i}_{t}^{u}=i_{t+1}^{u})
\log(p_{D}(\hat{\mathbf{h}}_{t}^{u})) \\ 
&- \mathbbm{1}(\hat{i}_{t}^{u}\neq i_{t+1}^{u})\log(1 - p_{D}(\hat{\mathbf{h}}_{t}^{u})).
\end{split}
\end{equation}

\begin{table*}[htb]
  \caption{Overall performance comparisons. 
  The last row is the relative
  improvements compared between the the best (bold) and the second best (underlined) scores.
}
  \label{tab:main-results-vt}
  \scalebox{0.9}{\setlength{\tabcolsep}{0.85mm}{
  \begin{tabular}{l|cccc|cccc|cccc|cccc}
    \toprule
    \multicolumn{1}{c|}{\multirow{2}{*}{SR Model}} & 
    \multicolumn{4}{c|}{\multirow{1}{*}{Yelp}} &
    \multicolumn{4}{c|}{\multirow{1}{*}{Beauty}} &
    \multicolumn{4}{c|}{\multirow{1}{*}{Sports}} &
     \multicolumn{4}{c}{\multirow{1}{*}{Toys}} 
      \\
      \cline{2-17}
      & 
    \multicolumn{2}{c}{\multirow{1}{*}{HR}}&
    \multicolumn{2}{c|}{\multirow{1}{*}{NDCG}}
     &
    \multicolumn{2}{c}{\multirow{1}{*}{HR}}&
    \multicolumn{2}{c|}{\multirow{1}{*}{NDCG}}
    &  
    \multicolumn{2}{c}{\multirow{1}{*}{HR}}&
    \multicolumn{2}{c|}{\multirow{1}{*}{NDCG}} &
    \multicolumn{2}{c}{\multirow{1}{*}{HR}}&
    \multicolumn{2}{c}{\multirow{1}{*}{NDCG}}
    \\
     &  
    @5 & @10 &
    @5 & @10 &
    @5 & @10 &
    @5 & @10 &
    @5 & @10 &
    @5 & @10 &
    @5 & @10 &
    @5 & @10  \\
    \midrule
    PopRec & 0.0057&0.0099&0.0037&0.0051 & 0.0080&0.0152&0.0044&0.0068 & 0.0056&0.0094&0.0041&0.0053 & 0.0066&0.0113&0.0046&0.0061 \\
    ${\text{Seq-BPR}}$ & 0.0139 & 0.0236 & 0.0087 & 0.0118 & 0.0195 & 0.0361 & 0.0113 & 0.0166 & 0.0145 & 0.0228 & 0.0091 & 0.0118 & 0.0242 & 0.0415 & 0.0148 & 0.0204  \\
    Caser & 0.0142&0.0253&0.008&0.0129 & 0.0251&0.0347&0.0145&0.0176 & 0.0154&0.0194&0.0114&0.1424 & 0.0166&0.0270&0.0107&0.0141\\
    GRU4Rec & 0.0152&0.0263&0.0091&0.0134 & 0.0164&0.0283&0.0099&0.0137 & 0.0162&0.0204&0.0103&0.0110 & 0.0097&0.0176&0.0059&0.0084\\
    SASRec & 0.0172 & 0.0289 & 0.0107 & 0.0144 & 0.0384  & 0.0607  & 0.0249  & 0.0321 & 0.0206 &  0.032 & 0.0135 & 0.0172 & 0.0484 & 0.0696 & 0.0329 & 0.0398  \\
    BERT4Rec & 0.0196 & 0.0339  & 0.0121 & 0.0167 &  0.0351 & 0.0601 & 0.0219 & 0.0300 &  0.0217 &  0.0359 &  0.0143 &  0.0190 & 0.0412 & 0.0594 & 0.0297 & 0.0348 \\
    S$^3\text{-Rec}$ & \underline{0.0201} & \underline{0.0341} & \underline{0.0123} & \underline{0.0168} & 0.0387 & 0.0647 & 0.0244 & 0.0327 & 0.0251 & 0.0385 & 0.0161 & 0.0204 & 0.0443 & 0.0700 & 0.0294 & 0.0376 \\
    Seq2Seq & 0.0183 & 0.0304 & 0.0119 & 0.0162 & \underline{0.0410} & \underline{0.0689} & \underline{0.0261} & \underline{0.0358} & \underline{0.0255} & \underline{0.0392} & \underline{0.0170} & \underline{0.0212} & \underline{0.0502} & \underline{0.0721}  & \underline{0.0337} & \underline{0.0421}  \\
    \midrule
    \modelname (\textbf{ours}) & \textbf{0.0434} &\textbf{0.0593} &\textbf{0.0218} & \textbf{0.0270} & \textbf{0.0705} & \textbf{0.0968} & \textbf{0.0507} & \textbf{0.0591} & \textbf{0.0380} & \textbf{0.0548} & \textbf{0.0268} & \textbf{0.0322} &\textbf{ 0.0756} & \textbf{0.0997} & \textbf{0.0557} & \textbf{0.0634} \\
    \midrule
    Improv. (\%) & 137.16&95.07&83.19&66.67 & 71.95&40.49&94.25&65.08 & 49.02&39.80&57.65&51.89 & 50.60&38.28&65.28&50.59\\
  \bottomrule
\end{tabular}}}
\end{table*}

\subsection{Training and Inference}
During the training stage, we perform jointly training to minimize the following losses:

\begin{equation}
\label{eq:all}
\begin{split}
\mathcal{L} = \mathcal{L}_{NIP} (I, G) + \lambda \mathcal{L}_{Disc} (\hat{I}, D),
\end{split}
\end{equation}
where $\lambda$ controls the strength of the discriminator. 
The item embeddings in both generator and discriminator are shared. We also 
explore the potential of sharing the Transformer encoder parameters. See Section~\ref{sec:ablation}
for detailed comparisons. The training objective is different from a GAN-based approach
because the generator is trained with the NIP task following the maximum log-likelihood principle
instead of aiming at adversarial fool the discriminator. If the generator generates the true
target items, then these items are considered as next items rather than `fake' ones.
The generator will be thrown out in the inference stage, leaving the trained discriminator 
as the final SR model.


\section{Experimental Study}

In this section, we conduct extensive experiments
to answer following research questions (RQs):
\begin{itemize}
    \item \textbf{RQ1}: Is \modelname effective and efficient for sequential recommendations compared with
    other baselines?
    \item \textbf{RQ2}: How does each components of \modelname influences \modelname's performance?
    \item \textbf{RQ3}: What's the optimal sampling rate $\alpha$ and weights of discriminator $\lambda$?
\end{itemize}

\subsection{Experiment Setup}

\subsubsection{Data}

We conduct experiments on four datasets
with various data distributions:
\emph{Yelp}\footnote{https://www.yelp.com/dataset} 
is a dataset for business recommendation..
\emph{Sports}, \emph{Beauty}, and \emph{Toys}
are three datasets collected from Amazon in~\cite{mcauley2015image}.
We follow~\cite{zhou2020s3,ma2020disentangled,qiu2021memory} 
to prepare the datasets.
Specifically, we  discard  all users and items that have
fewer than 5 related interactions. 
For each user, we use her last interacted item for testing, 
the second from the last item for validation,
and the rest items for training.

\subsubsection{Evaluation Metrics}
We follow previous works~\cite{wang2019neural,krichene2020sampled} 
to rank the predictions over 
the whole item set without negative sampling. 
Performance is
evaluated on 
a variety of Top-K evaluation metrics:
\textit{Hit Ratio}$@k$ ($\mathrm{HR}@k$), 
and \textit{Normalized Discounted 
Cumulative Gain}$@k$ ($\mathrm{NDCG}@k$) where $k\in\{5, 10\}$.

\subsubsection{Baseline}
We compare our method with eight different baselines.
An item popularity-based method: PopRec;
four deep SR methods that trained with next-item prediction task:
Caser~\cite{tang2018personalized}, GRU4Rec~\cite{hidasi2015session}, SASRec~\cite{kang2018self}, and S$^3\text{-Rec}$~\cite{zhou2020s3};
Two deep SR methods that trained with other generative tasks:
BERT4Rec~\cite{sun2019bert4rec} and Seq2Seq~\cite{ma2020disentangled}.

\subsubsection{Parameter Setting}
The number of attention heads and layers for all Transformer-based
methods are tuned from $\{1, 2, 4\}$ and $\{1, 2, 3\}$, respectively.
The embedding size is searched from  $\{32, 64, 128\}$.
The number of latent factors in Seq2Seq is tuned from $\{1, 2, \dots,8\}$.
We fixed the batch size and maximum sequence length for all methods as 256 and 50, respectively.
We implement our method in PyTorch. 
The sample rate $\alpha$ and the strength of discriminator $\lambda$ 
are both tuned from $\{0.0, 0.1, \dots,1.0\}$.
We tune
hyper-parameters on the validation set, 
and stop training if validation performance does not improve for 40 epochs.
We evaluate the final performance of all methods on test set.

\subsection{Overall Comparison}

Table~\ref{tab:main-results-vt} shows the overall performance comparisons
on four datasets.
First of all, our method consistently performs best among all methods over all datasets.
The improvements range from 38.28\% to 137.16\% in HR and NDCG, 
respectively, 
compared with the best baseline method. 
This observation demonstrates the effectiveness of \modelname and shows
that an SR trained with the proposed training framework 
can learn a more accurate
sequence and item representations.
Seq2Seq achieves the second-best results in three out of four
datasets, verifying the efficacy of leveraging multiple future items
to perform 
additional self-supervised learning.
S$^3\text{-Rec}$ fuses item attributes via contrastive self-supervised learning in pre-training stage and 
achieves the second-best results on Yelp. It verifies
that contrastive self-supervised learning is an effective way of 
leveraging additional item attributes in SR.
BERT4Rec replaces the next-item task with a mask-item prediction task
to leverage contextual information in the sequence and
perform better than SASRec in Yelp and Sports.
It shows that considering contextual information can benefit model learning.
However, BERT4Rec fails to outperform SASRec on Beauty and Toys,
which indicates that the mask-item prediction task may not align 
with the goal of recommendation (next-item prediction) well.
SASRec that encodes sequence with Transformer
outperforms CNN and RNN based approaches (Caser and GRU4Rec),
demonstrating the effectiveness of Transformer 
for capturing user dynamic changed behaviors.
The static method PopRec performs 
worst on all datasets as it ignores the sequential dynamics of user 
behaviors.

\begin{figure}[htb]
  \centering
  \includegraphics[width=0.9\linewidth]{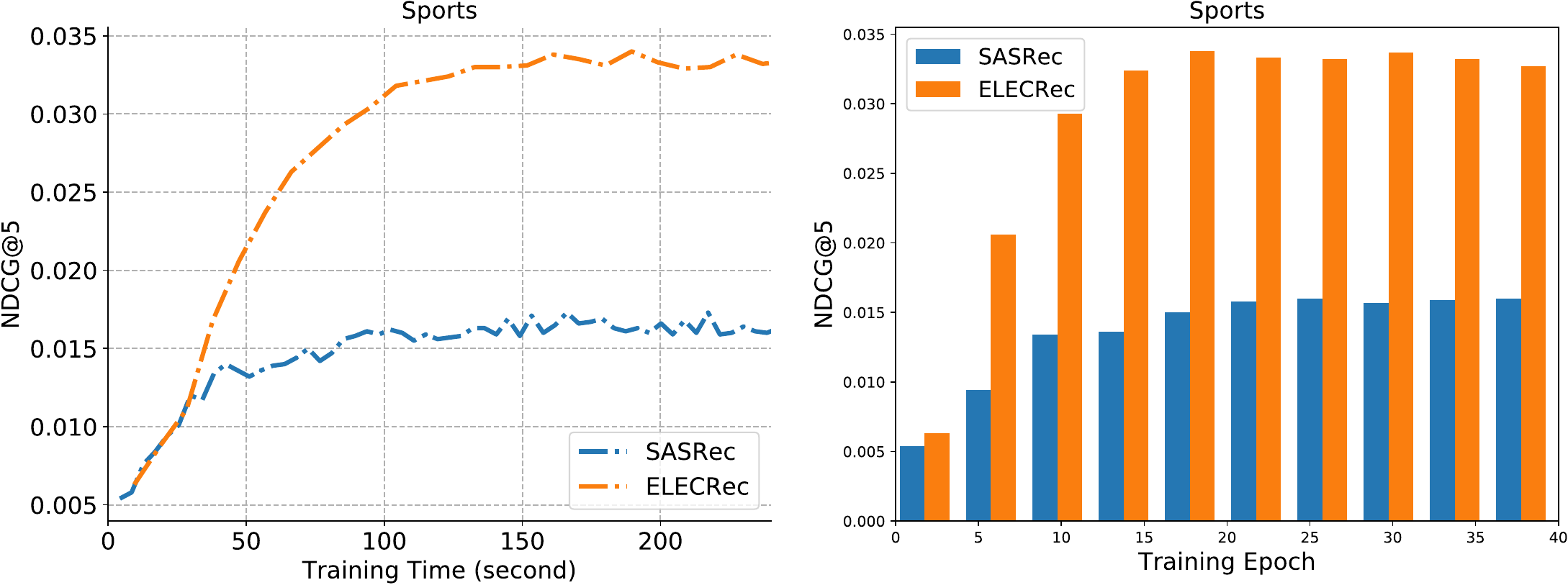}
  \caption{Efficiency comparison (Performance on validation set). 
  \modelname that trained with discriminator can 
  consistently outperform a typical SR model that trained with NIP task (e.g., SASRec)
  under any training time.
 }
  \label{fig:efficiency-study}
\end{figure}

We also study the proposed method's efficiency on the Sports 
dataset compared with the typical SR model SASRec trained with a NIP task. 
Figure~\ref{fig:efficiency-study} shows
the performance on validation set over training epochs and training time.
We can observe that, firstly, \modelname can outperform
SASRec under any training time and training epochs. The faster converge rate of \modelname 
demonstrates the efficiency of the proposed training strategy.
Besides, \modelname can converage to much higher performance than SASRec.
This phenomenon shows the crucial of distinguishing plausible items
for improving the accuracy of users' preferences towards large vocabulary items.

\begin{table}[htb]
\caption{Ablation study of \modelname on Yelp and Beauty (HR@5 and NDCG@5). }
\label{tab:ablation-study}
\scalebox{1.0}{\setlength{\tabcolsep}{1.4mm}{
\begin{tabular}{l|cc|cc}
\toprule
\multirow{2}{*}{Model}  &\multicolumn{2}{c}{\multirow{1}{*}{Yelp}}&
    \multicolumn{2}{c}{\multirow{1}{*}{Beauty}}\\
    \cline{2-5}
    &  HR & NDCG &  HR & NDCG \\
\hline
(A) $\modelname_{ES}$  & \textbf{0.0434} &\textbf{0.0218} & 0.0682&0.0487\\
(B) $\modelname_{FS}$ & 0.0334&0.0168 & \textbf{0.0705} & \textbf{0.0507} \\
(C) Generator only & 0.0248 &0.0157 &  0.0685 & 0.0473\\
(D) SASRec & 0.0172   & 0.0107  & 0.0384 & 0.0249\\

\bottomrule
\end{tabular}}}
\end{table}

\subsection{Ablation Study}
\label{sec:ablation}
We conduct a detailed ablation study on Yelp and Beauty to verify
the effectiveness of each component and report the results
in Table~\ref{tab:ablation-study}. 
(A) $\modelname_{ES}$ and (B) $\modelname_{FS}$ are two types of \modelname that either sharing the weights of item embeddings or
with additional Transformer parameters
between generator and 
discriminator, respectively.
(C) only trains the generator as the final SR model.
(D) is SASRec for comparison. The key difference between (C) and (D) is that (D) trains the generator via a 
sequential binary cross-entropy (BCE) loss while (C) trains the generator 
via a sequential multi-classes cross-entropy (Softmax) loss so that all item embeddings
are updated densely during training. From (A), (B), and (C) we can see that
the discriminator does benefits the model learning. Intuitively,
the discriminator tries to update the embeddings of plausible items, 
either sampled from the generator or 
present in the original sequence so that the SR model can distinguish them more easily.
From (C) and (D) we can also see that a model trained with Softmax can achieves better performance
than its binarized approximation (sequential BCE).




\begin{figure}[htb]
  \centering
  \includegraphics[width=1.0\linewidth]{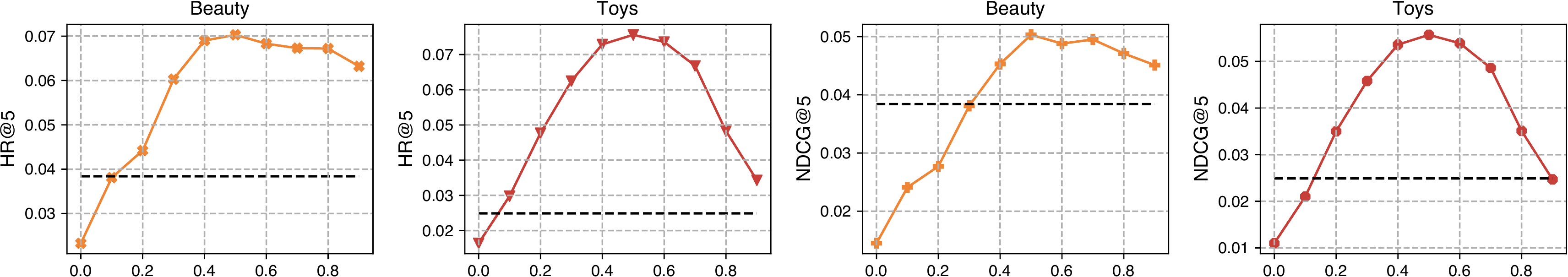}
  \caption{Performance w.r.t. the sampling rate $\alpha$. The black dash line
  is the performance of SASRec for comparison.
 }
  \label{fig:impact-alpha}
\end{figure}

\subsection{Influence of the sampling rate $\alpha$ and discriminator weight $\lambda$}
\begin{figure}[htb]
  \centering
  \includegraphics[width=1.0\linewidth]{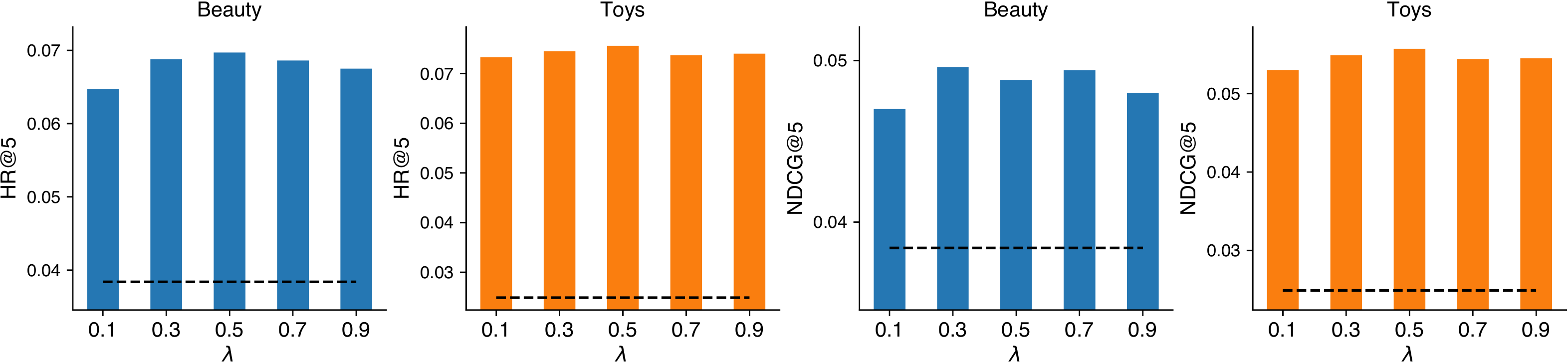}
  \caption{Performance w.r.t. discriminator weight $\lambda$. 
  The black dash line is the performance of SASRec for comparison.
 }
  \label{fig:impact-lambda}
\end{figure}

\modelname introduces two hyper-parameters:
the sampling rate $\alpha$ and the weights to
control the strength of the jointly
trained discriminator.
We conduct experiments on Beauty and Toys and show
the influence of these two hyper-parameters
in Figure~\ref{fig:impact-alpha} and Figure~\ref{fig:impact-lambda},
respectively.
From Figure~\ref{fig:impact-alpha} we can see that
$\alpha=0.5$ is the optimal value on Beauty and Toys.
When $\alpha$ is too large, most of items in the sequence are replaced by the generator
and viewed as negative class.
When $\alpha$ is too small, most of items are from original input
and are viewed as positive class. 
The brought of data imbalance issue can affects learning thus deteriorates performance.
From Figure~\ref{fig:impact-lambda} we can observe
$\lambda=0.5$ gives the best performance on Beauty and Toys in HR@5. 
While model with $\lambda=0.3$ perform best on Beauty in terms of NDCG@5.
In general, the $\lambda$ less than 1.0 benefits the model learning most.
This phenomenon indicates that the discriminator's job 
is to identify the plausible items sampled from the generator, 
which are a small group of items. 
In comparison, the generator takes more responsibility 
for training the whole item embeddings, 
thus requires larger weights to update the parameters.

\section{Conclusion \& Limitations}

In this work, 
we propose to train sequential 
recommenders as discriminators instead of 
generators so that 
the user behavior sequence and item representations 
are more accurate. 
We propose to generate high-quality training samples
for the discriminator via a jointly trained generator
so that the discriminator can
steadily improve its ability to discriminate 
the true item correlations.
Extensive experiments on four datasets with eight baseline methods 
demonstrate the effectiveness of our propose training framework. 
Detailed ablation study and analysis also 
shows the superiority of \modelname that trained via our propose training scheme.
A limitation of our work is that
the model performance needs to be carefully tuned 
and optimal hyper-parameters vary over datasets.


\bibliographystyle{ACM-Reference-Format}
\balance
\bibliography{sample}
\newpage
\appendix

\end{document}